\documentclass[10pt,journal,compsoc]{IEEEtran}

\usepackage{scalerel}
\usepackage{tikz}
\usetikzlibrary{svg.path}
\usepackage{epstopdf}
\usepackage{multirow}

\usepackage{amsmath,amssymb,amsfonts}
\usepackage{algorithmic}
\usepackage{graphicx}
\usepackage{subfigure}
\usepackage{booktabs}  
\usepackage{lipsum} 
\usepackage{textcomp}
\usepackage{wrapfig}
\usepackage{multirow}
\usepackage{color}
\usepackage{hhline}
\definecolor{RedColor}{rgb}{1,0,0}
\definecolor{BlueColor}{rgb}{0,0,1}
\definecolor{GreenColor}{rgb}{0,1,0}
\usepackage[colorlinks, linkcolor=red,  anchorcolor=blue,     citecolor=green            ]{hyperref}
\definecolor{orcidlogocol}{HTML}{A6CE39}
\tikzset{
	orcidlogo/.pic={
		\fill[orcidlogocol] svg{M256,128c0,70.7-57.3,128-128,128C57.3,256,0,198.7,0,128C0,57.3,57.3,0,128,0C198.7,0,256,57.3,256,128z};
		\fill[white] svg{M86.3,186.2H70.9V79.1h15.4v48.4V186.2z}
		svg{M108.9,79.1h41.6c39.6,0,57,28.3,57,53.6c0,27.5-21.5,53.6-56.8,53.6h-41.8V79.1z M124.3,172.4h24.5c34.9,0,42.9-26.5,42.9-39.7c0-21.5-13.7-39.7-43.7-39.7h-23.7V172.4z}
		svg{M88.7,56.8c0,5.5-4.5,10.1-10.1,10.1c-5.6,0-10.1-4.6-10.1-10.1c0-5.6,4.5-10.1,10.1-10.1C84.2,46.7,88.7,51.3,88.7,56.8z};
	}
}

\newcommand\orcidicon[1]{\href{https://orcid.org/#1}{\mbox{\scalerel*{
				\begin{tikzpicture}[yscale=-1,transform shape]
					\pic{orcidlogo};
				\end{tikzpicture}
			}{|}}}}

\usepackage{hyperref} 

\usepackage{graphicx}
\usepackage{float} 
\usepackage{subfigure}
\usepackage{color}
\usepackage{amsmath}

\ifCLASSOPTIONcompsoc
  \usepackage[nocompress]{cite}
\else
  \usepackage{cite}
\fi
\ifCLASSINFOpdf
\else
\fi
\begin{document}
%
\title{Working memory inspired hierarchical video decomposition with transformative representations}
%
%
%
%

\author{Binjie Qin*\orcidicon{0000-0001-7445-1582},~\IEEEmembership{Member,~IEEE}, Haohao Mao\orcidicon{0000-0001-9730-6062}, Ruipeng Zhang\orcidicon{0000-0002-5378-4086}, Yueqi Zhu, Song Ding, Xu Chen\orcidicon{0000-0002-4870-2381}
\IEEEcompsocitemizethanks{\IEEEcompsocthanksitem Binjie Qin, Haohao Mao and Ruipeng Zhang are with the School of Biomedical Engineering, Shanghai Jiao Tong University, Shanghai 200240, China. E-mail: E-mail: bjqin@sjtu.edu.cn 
\IEEEcompsocthanksitem Yueqi Zhu is with the Department of Radiology, Shanghai Jiao Tong University Affiliated Sixth People's Hospital, Shanghai Jiao Tong University, 600 Yi Shan Road, Shanghai 200233, China.
\IEEEcompsocthanksitem Song Ding is with the Department of Cardiology, Ren Ji Hospital, School of Medicine, Shanghai Jiao Tong University, Shanghai 200127, China. 
\IEEEcompsocthanksitem Xu Chen is with the Center for Advanced Neuroimaging, UC Riverside, 900 University Avenue, Riverside, CA 92521, USA.}

\thanks{The manuscript was received on April 14, 2021, and accepted on October 27, 2018. This work was partially supported by the Science and Technology Commission of Shanghai Municipality (19dz1200500, 19411951507), the National Natural Science Foundation of China (61271320, 82070477), Shanghai ShenKang Hospital Development Center (SHDC12019X12), and the Interdisciplinary Program of Shanghai Jiao Tong University (ZH2018ZDA19, YG2021QN122, YG2021QN99). (Corresponding author: Binjie Qin.)}}

%
%

\markboth{Journal of \LaTeX\ Class Files,~Vol.~14, No.~8, August~2015}%
{Shell \MakeLowercase{\textit{et al.}}: Bare Advanced Demo of IEEEtran.cls for IEEE Computer Society Journals}
%



\IEEEtitleabstractindextext{%
\begin{abstract}
Video decomposition is very important to extract moving foreground objects from complex backgrounds in computer vision, machine learning, and medical imaging, e.g., extracting moving contrast-filled vessels from the complex and noisy backgrounds of X-ray coronary angiography (XCA). However, the challenges caused by dynamic backgrounds, overlapping heterogeneous environments and complex noises still exist in video decomposition. To solve these challenges, this study is the first to introduce a flexible visual working memory model in video decomposition to provide interpretable and high-performance hierarchical deep learning architecture, integrating the transformative representations between sensory and control layers from the perspective of visual and cognitive neuroscience. Specifically, robust PCA unrolling networks acting as a structure-regularized sensor layer decompose XCA into sparse/low-rank structured representations to separate moving contrast-filled vessels from noisy and complex backgrounds. Then, patch recurrent convolutional LSTM networks with a backprojection superresolution module embody unstructured random representations of the control layer in working memory, recurrently projecting spatiotemporally decomposed nonlocal patches into orthogonal subspaces for heterogeneous vessel retrieval and interference suppression. This video decomposition architecture effectively restores the heterogeneous profiles of intensity and  geometry of moving objects against the complex background interferences. Experiments show that the proposed method significantly outperforms state-of-the-art methods in accurate moving contrast-filled vessel extraction with excellent flexibility and computational efficiency.
\end{abstract}

}

\maketitle

\IEEEdisplaynontitleabstractindextext

%
\IEEEpeerreviewmaketitle

\ifCLASSOPTIONcompsoc
\IEEEraisesectionheading{\section{Introduction}\label{sec:introduction}}
\else
\section{Introduction}
\label{sec:introduction}
\fi

%
%
%
%
\IEEEPARstart{V}{ideo} decomposition into foreground/background components is very important for moving object extraction in computer vision, machine learning, and medical imaging\cite{Chakraborty2021Intrinsic,shao2021hyper,jin2017extracting,ma2017automatic,xia2020vessel}. Simply subtracting a static background frame from the current frame may easily lead to incomplete foreground extraction due to the following immediate changes in real scenarios: motion variations of dynamic background\cite{yong2018robust} and camera, illumination and intensity changes in background/foreground components, and complex noises occurring in low-light images. Among all real scenarios for foreground/background separation, separating vessels from dynamic and complex backgrounds in X-ray coronary angiography (XCA) is the most representative application that covers all challenging problems. Specifically, XCA via low-dose X-ray imaging projects 3D objects onto a 2D plane to image blood vessels in the diagnosis and treatment of cardiovascular diseases (CVDs), such that XCA vessels have low-contrast structures that overlap with complex backgrounds with their accompanying motion interferences and vessel-like artefacts as well as signal-dependent mixed Poisson-Gaussian noises\cite{Irrera2016flexible,zhao2019texture}. In addition, the blood flow in CVDs is usually laminar and dependent on the vessel radius, with its velocity profile over the longitudinal section being parabolic\cite{Bradley1985Blood}. The change in vessel curvature along with the development of fatty plaques that can narrow and/or clog blood vessels also contribute to blood flow reduction in CVDs. These factors lead to the high spatiotemporal heterogeneity of XCA vessels, which becomes more prominent in the XCA sequences acquired from different patients or imaging equipments. These challenges have motivated increasing efforts to accurately extract overlapping heterogeneous vessels from XCA sequences in recent years.  
\begin{figure*}[ht]
	\centering 
	\includegraphics[width=0.61\textwidth]{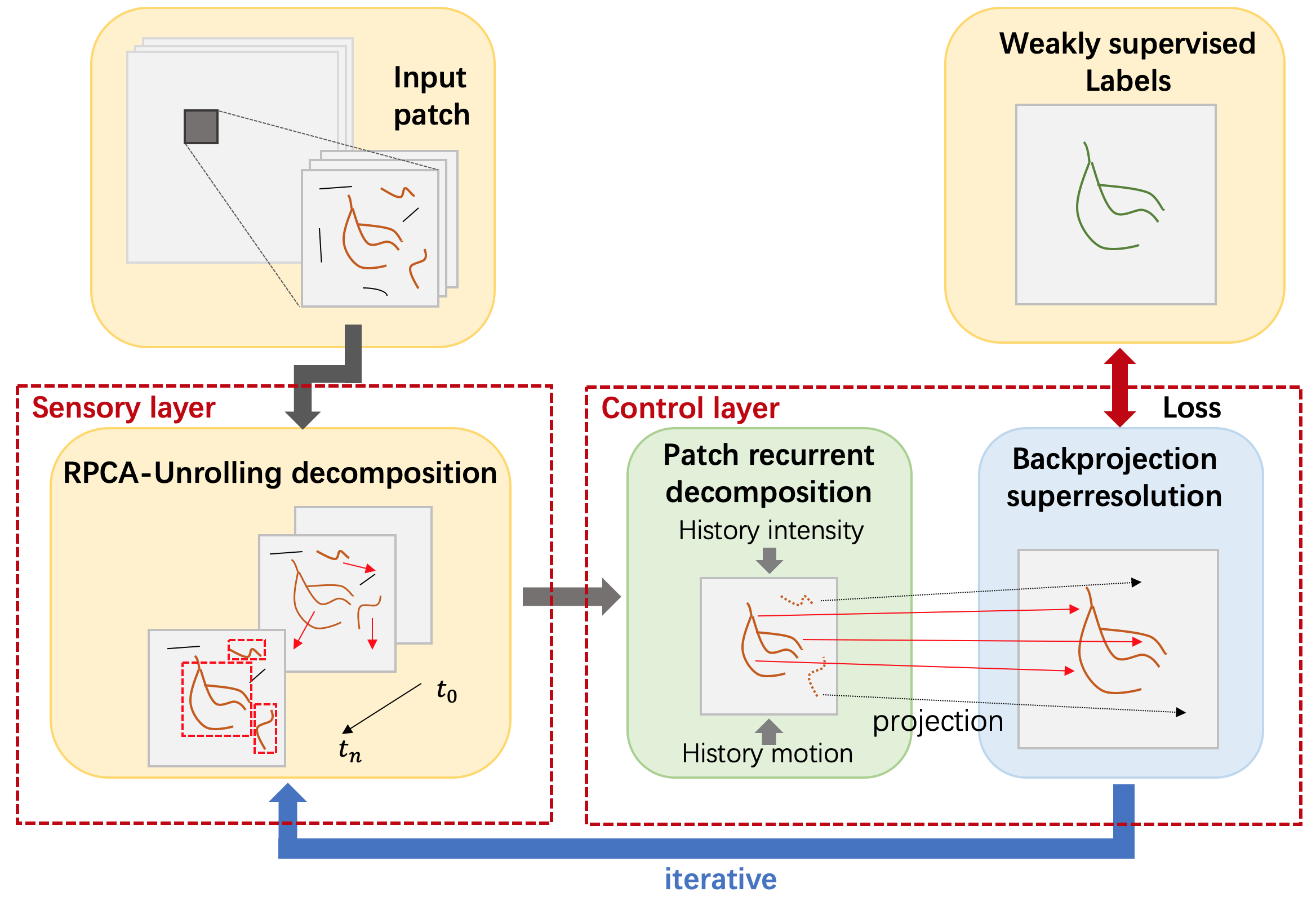}
	\caption{Working memory-inspired dual-stage patch recurrent unrolling architecture for video decomposition, which mainly includes the sensor layer of RPCA unrolling networks and the control layer of patch recurrent backprojection superresolution module, smoothly transforming the structured representations of global foreground/background decomposition into the unstructured random representations of patch recurrent orthogonal decomposition.} 
	\label{Fig1} 
\end{figure*}

In fact, few recent studies have been conducted on developing moving contrast-filled vessel extraction algorithms\cite{moccia2018blood,jia2021learning}, which can be mainly categorized into four types: vessel enhancement \cite{wan2018automated}, deformable model\cite{ge2019two}, vessel tracking\cite{fang2019greedy,yang2020vessel}, and machine learning \cite{yang2020vessel}. We refer interested readers to recent comprehensive reviews on XCA vessel extraction\cite{qin2022extracting,Qin2022RPCAUNet}. However, most vessel extraction algorithms are built upon grey value or tubular feature representation, which overlap with the interferences of complex noises and dynamic background artefacts. Recently, assuming $\mathbf{D}=\mathbf{L}+\mathbf{S}$, where $\mathbf{D},\mathbf{L},\mathbf{S}\in \mathbb{R}^{m\times n}$ are the original video sequence, low-rank backgrounds, and sparsely distributed foreground objects, respectively, robust principal component analysis (RPCA)\cite{Candes2011Robust,bouwmans2018applications} has proven to successfully separate moving contrast-filled vessels from complex and dynamic backgrounds\cite{jin2017extracting,ma2017automatic,jin2018low,song2020spatio,xia2020vessel}. When only a subset of the entries of $\mathbf{D}$ is observed, RPCA becomes the robust low-rank matrix (or tensor) completion that has been explored to complete the background layer of the XCA sequence for accurate vessel extraction\cite{qin2019accurate}. However, RPCA methods still include some noisy artefacts and require iterative numerical models that are prohibitively costly for clinical applications. Therefore, a convolutional robust PCA (CORONA)\cite{solomon2019deep} unrolls the RPCA into a deep neural network to greatly improve the time efficiency, while the extracted vessels still contain many noisy artefacts. 

One challenge of moving object extraction under noisy and dynamic backgrounds is how to deal with occluded or overlapped objects with motion interferences and noisy artefacts. In XCA imaging, low-contrast blood vessels of all radii are seriously overlapped by dynamic structures and some vessel-like artefacts in noisy backgrounds. The large variability of the overlapping structures in terms of their shape, appearance and motion profile introduces exponential complexity in the video data distribution that is highly elusive to exhaustive representation in finite training data. Recent works\cite{Shao2022Exploiting,Sultana2021Background,Bouwmans2019Deep} have shown that foreground/background decomposition-based deep vision systems for object extraction in overlapping areas are not as robust as human vision in separating multiple overlapping objects, let alone most supervised deep learning approaches in indiscriminately learning the structure of all image pixels using labelled data. Moreover, this limitation has been intensified by the variations in heterogeneous environments even when deep networks exposed to a large amount of partial occlusion during training have exploited attention modules\cite{Shao2022Exploiting,Sultana2021Background} with efficient loss terms and an effective generator network\cite{Sultana2021Background} to guide foreground segmentation.  

While robustness to overlapping heterogeneity is crucial, safety-critical applications also require AI systems to provide human-interpretable explanations of their predictions in accordance with prior knowledge. Such interpretability can potentially support the scientific understanding of the human vision process to advance high-performance AI systems. In fact, visual working memory\cite{buschman2021balancing,Robert2021Working,Xie2022Geometry} serves as a unitary cognitive system over short- and long-term memories in human vision and has a key cognitive capacity to track relevant information from a sequence of events. This capacity is desired in engineering tasks for processing sequential data and easily recognizing overlapping heterogeneous vessels from XCA sequences. Specifically, a flexible working memory model\cite{buschman2021balancing} is regarded as a cognitive computation system to use distributed and factorized representations in different partially overlapped feature spaces for distributively implementing cognitive functions as matrix multiplication\cite{Xie2022Geometry}. By acting as compatible models that implemented vector symbolic architectures and tensor-product representations for video sequences, a sensory layer with structured representation (or encoding) distributively perceived the sparsely diffused features from the interferences occurring in the overlapped feature spaces, while a shared unstructured control layer with random recurrent connections, with balanced excitation and inhibition for each neuron as the only constraint, recurrently projected the perceived representation into discrete states of orthogonally decomposed subspaces for recognizing sequential items from the overlapping interferences.

Inspired by the flexible working memory model, this work proposes a dual-stage video decomposition architecture for XCA vessel extraction by implementing a multiscale RPCA unrolling network with a patch recurrent backprojection (msRPCA-PBNet) module (see Fig. \ref{Fig1}): a sensory layer that inspires the implementation of RPCA unrolling globally decomposes the foreground/backgrounds via structured representation of XCA sequences; after inputting structural intensity and motion information of foreground vessels, a patch recurrent convolutional long short-term memory (CLSTM)\cite{shi2015convolutional} combined with backprojection\cite{haris2021deep} superresolution and upsampling embodies the random representation of the control layer to recurrently project these foreground candidates into spatiotemporally decomposed nonlocal patches, while the topological structure of candidate input is maintained and adaptively attracted into discrete contextually predicted values\cite{panichello2019error}. The main contribution of this work is threefold:

1) We propose a working memory-inspired dual-stage patch recurrent unrolling architecture (WMIDPRUA) for video decomposition in a hierarchical and interpretable way to extract moving contrast-filled vessels from XCA video with superb accuracy and computation efficiency. To the best of our knowledge, this is the first study to use working memory theory to design a deep video decomposition architecture with better interpretability and efficiency for solving the challenging foreground/background separation problem. The proposed msRPCA-PBNet with the underlying transformative representations of WMIDPRUA smoothly transforms the structured representations of RPCA-unfolded global foreground/background decomposition into the unstructured random representations of nonlocal patch recurrent decomposition, achieving heterogeneous vessel superresolution and interference suppression via patch recurrent backprojection\cite{haris2021deep} in spatiotemporally orthogonal subspaces. 


2) We integrate a CLSTM-based feature selection\cite{Qin2022RPCAUNet} into random backprojection\cite{haris2021deep} superresolution to introduce a spatiotemporally decomposed nonlocal patches in orthogonal subspaces for selecting spatial features and investigating temporal dynamics in vessel heterogeneity recovery and interference suppression. The patch recurrent CLSTM mechanism leads to fewer vessel representations that need to be learned by the networks, as increasing the sparsity of neural representations in the orthogonal subspaces can reduce overlapping interferences in working memory. Moreover, due to continuous deformation interferences and mixed Poisson-Gaussian noises being approximated as local movements and Gaussian noises in each patch, respectively, the proposed patch recurrent backprojection facilitates the clustering of nonlocally similar patches to remove complex noise and motion interferences, while the memory cell with random backprojection can enable better vessel heterogeneity identification from the overlapping vessel-like background artefacts.

3) We extend the WMIDPRUA with representation transformation into a hierarchical multiscale architecture, such that the hierarchical WMIDPRUA can achieve transformative representation hierarchy at different scales of the input video patches from XCA sequences, where short- and long-range local/nonlocal spatiotemporal correlation between the recurrent patches can be used to distinguish between heterogeneous vessels and noisy background artefacts. The experiments show that the multiscale WMIDPRUA architecture is beneficial to further eliminate background interferences. 	

The remainder of this paper is organized as follows. Section \ref{sec:Related Work} reviews the related work in detail. Section \ref{sec:method} introduces the proposed msRPCA-PBNet built on WMIDPRUA for XCA video decomposition. Section \ref{sec:Experimental results} presents the experimental results. Section \ref{sec:Conclusion} summarizes the conclusion and discussion.

\section{Related Work}\label{sec:Related Work}
\subsection{RPCA-based Foreground/Background Separation}
By robustly learning the intrinsic low-dimensional subspace from high-dimensional data, RPCA is a commonly used foreground/background separation technique for moving foreground object extraction in a video sequence taken by a static camera. Bouwmans \textit{et al.}\cite{bouwmans2017decomposition} provided a comprehensive survey of RPCA for foreground/background separation, which differs from decomposition regularization, noise modelling, the loss function, the minimization problem and the solvers used. Recently, RPCA has mainly been developed into the following distinct research lines by investigation of its adequacy for the application of foreground/background separation: 

\textbf{Integrating spatiotemporally-regularized representations} for low-rank backgrounds and sparse foregrounds into the RPCA loss function can ensure the uniqueness of the decomposition solution with high foreground/background separation performance. Instead of applying the $l_{1}$-norm\cite{jin2017extracting,song2020spatio,xia2020vessel} and $l_{1/2}$-norm\cite{Zhu2018L1/2,Tom2021Three} to globally select sparse foreground features, recent studies have investigated the structured sparsity over groups of spatiotemporally neighbouring pixels, such as graph-based regularization\cite{Javed2019Moving}, superpixel-based clustering\cite{xia2020vessel}, Gaussian mixture distribution\cite{zhao2014robust,Ran2021Anomaly}, Markov random field constraint\cite{Zhou2013Moving}, tree-structured regularization\cite{Ebadi2018Foreground}, kinematic
regularization\cite{Eltantawy2019Accelerated}, and total variation norm\cite{Cao2016Total,jin2017extracting,Tom2021Three}, while alternative strategies have used higher-order tensor instead of matrix representation of video data for tensor RPCA (or robust tensor decomposition)\cite{Lu2020Tensor,Gao2021Enhanced} by specifying different tensor rank definitions and corresponding low-rank regularizations to explore an intrinsic spatiotemporal structure underlying multidimensional tensor data. 

However, most regularization approaches achieve global low-rankness and sparsity constraints on the RPCA composite objective function using the weighted sum, such that the possible local/nonlocal interactions and associations between heterogeneous subpopulations from the foregrounds or backgrounds, especially in the heterogeneous mixture of low-rank\cite{Chen2021Learning} and sparse structures, are usually neglected for video data analysis. In XCA vessel extraction, the mixture structures reflect complex spatiotemporal correlations that are spatiotemporally embedded into the noisy backgrounds, e.g., both low-rank and non-low-rank vessel-like artefacts as well as the distal vessels and vascular walls that show slow-moving contrast agents being gradually dissipated from the foregrounds to the backgrounds. Accurately restoring these distal vessels and vascular walls is especially important and particularly difficult for quantitative microcirculation analysis.

\textbf{Ensuring robustness to interferences} in RPCA-based applications is often perturbed by complex noises\cite{zhao2014robust} such as signal-dependent mixed Poisson-Gaussian noises in X-ray imaging. To robustly mitigate the complex noise disturbances, some stable decomposition formulations were developed into three-term decomposition that includes a noise component\cite{Oreifej2013Simul}, expressing a single independent identically distributed (i.i.d.) distribution, such as Gaussian and Laplacian \cite{yong2018robust,shao2021hyper}, or even sparser components\cite{Tom2021Three}, while recent works favour mixed noise modelling, such as a mixture of Gaussian (MoG)\cite{zhao2014robust,yong2018robust} and an information-theoretic learning strategy\cite{Guo2018GoDec+,Gao2022Robust}, to quantify the noise perturbation. However, these methods cannot model the spatiotemporally distributed signal-dependent noises in the XCA-like heterogeneous environments related to patient and device variability or non-identically distributed data. 

In addition to choosing the noise model, defining background and camera motions, their representation and the RPCA loss function used for modelling and updating the low-rank subspace are particularly impactful. Traditionally, modelling the background/camera motions with rigid or affine transformations\cite{yong2018robust,Eltantawy2019Accelerated,Moore2019Panoramic} is apparently impracticable for modelling the large local deformations of dynamic backgrounds in XCA imaging. 

\textbf{Implementing fast RPCA} for video decomposition is required to address the concerns of computational cost and sensitivity that result from the standard batch-based SVD calculation in each iteration of rank constrained optimization for low-rank subspace updating. Recently, incremental RPCA based on the fixed-rank constrained nonconvex approach has been greatly developed for tracking the gradually changing low-rank subspace. Recursive projected compressive sensing\cite{Qiu2014Recursive,Narayanamurthy2020Fast} projects the background's null space into a video frame to nullify the background while retaining the moving object, which can adapt to the dynamic background and camera motion by updating the subspace with faster rank-$r$ projection-SVD. Grassmannian rank-one update subspace estimation\cite{He2012Incremental,Hauberg2016Scalable,Chakraborty2021Intrinsic} identifies the subspace as a point on the Grassmannian manifold, the set of all linear subspaces of $ {\mathbb{R}}^{{n}} $ of fixed $r$-dimension. Chakraborty \textit{et al.}\cite{Chakraborty2021Intrinsic} further allow for fast computing of principal linear subspaces in finite and infinite dimensional reproducing kernel Hilbert spaces as well as for computing the principal components as the intrinsic average based on all Grassmannian subspaces. Thanh \textit{et al.} \cite{Thanh2021Robust} build upon parallel estimation and tracking by recursive least squares (PETRELS)\cite{Chi2013PETRELS} to detect and reject outliers via an alternating direction method of multipliers (ADMM) solver in a more effective way with an improved PETRELS for updating the underlying subspace. Low-rank matrix factorization\cite{Chi2019Nonconvex,yong2018robust} assumes that the rank-$r$ of $\mathbf{L}$ is known or initialized and then factorizes the $\mathbf{L}$ into the multiplication of two much smaller factored matrices for computational efficiency. Incremental principal component pursuit\cite{Rodriguez2016} exploits incremental and rank-one modifications for thin SVD in updating a low-rank background. However, these approaches face a clear limitation in batch initialization for the low-rank background and/or its rank estimation, which is too varying and sensitive to be exactly known a priori in noisy heterogeneous environments. These methods could fail if the largely distributed XCA vessels overlap with large portions of the complex background that exhibits complex motion and noise disturbances with vessel-like artefacts.

In recent years, randomized SVD algorithms \cite{Halko2011Finding,Oh2018Fast,Martinsson2020Randomized} have proven their outperformance over classical deterministic methods in the low-rank approximation of streaming data with limited working storage and minimum data communication. By exploiting low-rank approximation using random sampling to derive a smaller subspace that is aligned with the range of the input high-dimensional matrix, the key idea of the randomized method is to extract an approximate basis for the range of the input matrix from its sampled matrix. This finding of a subspace that captures much of the action of a matrix is called the rangefinder problem in randomized numerical linear algebra\cite{Martinsson2020Randomized}. Subsequently, randomized methods performed the deterministic decomposition (i.e., SVD) method on the smaller sampled subspace and then projected the decomposed results back to form the full decomposition for reducing the costs of computation, communication, and storage. Randomized algorithms typically use nonuniform sampling to select a certain set of row and column vectors from the target matrix, which can achieve an important sampling selection with lower overhead and higher accuracy compared with that of the uniform sampling method. Coupled with large data matrix partition schemes and a partial (or truncated) SVD of a small matrix, randomized SVD algorithms can be implemented in parallel on graphics processing units (GPUs) with the capability of fast matrix multiplications and random number generations to achieve further acceleration\cite{Lu2020Reducing,Struski2021Efficient}. Nevertheless, the computational bottleneck restricting real-time performance still exists in the CPU-GPU transfer bandwidth and vector summation\cite{Lu2020Reducing,Struski2021Efficient} inherent in RPCA-based video decomposition.

\subsection{Interpretable Deep Algorithm Unrolling}
Recently, interpretable deep learning has been primarily classified into two approaches, i.e., ad hoc and post hoc interpretability\cite{Tjoa2021Survey,Zhang2021Survey}; the former actively designs the network architecture, while the latter passively explains trained neural networks. Although some strategies of post hoc interpretability emphasize analysing the learned features using different techniques, such as attention mechanisms learning the importance of multidimensional features\cite{Jiang2021Decomposition}, layerwise relevance propagation explaining motion relevance for activity recognition\cite{hiley2020explaining}, and hidden semantics visualizing the behaviour of hidden layers for video change detection\cite{jimaging4060078}, few studies in video decomposition for moving object extraction attempt to provide ad hoc interpretability of deep learning-based models. 

Deep algorithm unrolling has recently received increasing attention in model-based interpretable deep learning by transforming iterative algorithms into deep neural networks for efficiently solving various inverse problems in image/video processing and compressive sensing\cite{Monga2021Algorithm}. The definition of deep unrolling was proposed by Gregor and LeCun\cite{gregor2010learning}, who unrolled the iterative shrinkage/thresholding algorithm (ISTA) to solve the optimization problem for sparse coding and achieved a nearly 20-fold improvement in time efficiency. Recently, by providing the neural network interpretability of iterative sparse coding with fewer layers and faster convergence, the ISTA-based deep unrolling algorithm has achieved great success in solving inverse problems for biomedical imaging\cite{xiang2021fista}, exploiting multimodal side information for image superresolution\cite{Marivani2020Multimodal}, and implementing nonnegative matrix factorization for functional unit identification\cite{Woo2021deep}.

Regarding unrolling RPCA, Sprechmann \textit{et al.}\cite{Sprechmann2015Learning} proposed a learnable pursuit architecture for structured RPCA decomposition to unroll the iteration of proximal descent algorithms for faithful approximation of the RPCA solution. However, this approach is largely dependent on a nonconvex formulation in which the rank of the low-rank background component is assumed to be known \textit{a priori}, but it is too varying to be estimated in real applications such as in XCA imaging. To overcome the heavy computation of RPCA, Solomon \textit{et al.}\cite{solomon2019deep} proposed convolutional robust PCA to unroll the ISTA for automatically separating vessels and background structures in ultrasound videos. Thanthrige \textit{et al.}\cite{thanthrige2022deep} proposed the reweighted $l_{1}$-norm and reweighted nuclear norm for RPCA regularization in unrolling the iterative algorithm of ADMM to improve the accuracy and convergence of recovering the low-rank and sparse components in material defect detection. Cai \textit{et al.}\cite{Cai2021Learned} proposed scalable and learnable feedforward recurrent-mixed neural networks using a simple formula and differentiable operators to avoid singular value thresholding of SVD during both training and inference for high-dimensional RPCA unrolling. However, the rank of the underlying low-rank matrix must be estimated as the input of RPCA unrolling. Unfortunately, these methods cannot overcome the complex interferences from signal-dependent mixed noises and dynamic background motions in heterogeneous environments. 

We have proposed RPCA-UNet\cite{Qin2022RPCAUNet} with a CLSTM-based feature selection mechanism to improve patch-wise vessel superresolution performance from noisy RPCA-unfolded results of XCA sequences. However, by selecting the vessel candidates from the structured representations of RPCA-unfolded results, the proposed RPCA-UNet is still unable to accurately restore continuous heterogeneity of XCA vessels while removing block and blur effects as well as residual interferences in XCA vessel extraction. Generally, without using memory-based smooth representation transformation to balance flexibility and interference in tracking a sequence of events, the power of deep algorithm unrolling networks does not seem to have been fully brought to bear on the challenging problem of high-performance architecture and its ad hoc interpretability for deep video decomposition. To solve this problem, an attempt at working memory inspired transformative representations is conducted in the proposed mcRPCA-PBNet to achieve ad hoc interpretability and computational efficiency of video decomposition with challenging XCA data.

\subsection{Working Memory Inspired Deep Learning}
Working memory is a capacity-limited but flexible cognition system to track a sequence of events using distributed representations and including perceptual as well as semantic information associated with the stimulus input and activated from knowledge/long-term memory\cite{buschman2021balancing,Robert2021Working}. By imitating working memory in maintaining the sequential information over time, existing recurrent neural networks (RNNs) that cyclically update their current state based on past states and current input data have incorporated an LSTM\cite{hochreiter1997long,Greff2017LSTM,Yu2019Review} module with or without convolutional structures\cite{shi2015convolutional} into recurrent architecture, including LSTM-dominated and integrated LSTM networks\cite{Yu2019Review} for various complicated reasoning and inference tasks related to sequential data. Specifically, by optimizing the connections of the inner LSTM cells for the performance enhancement of LSTM-dominated networks, adding learnable nonlinear state-to-gate memory connections performs noticeably better than the vanilla LSTM for various tasks with longer sequences\cite{LANDI2021334}, while conducting convolutional operation on the two input-to-state/state-to-state transitions and on the previous outputs/current input of the LSTM can integrate long-term temporal dependence with time-frequency characteristics\cite{Ma2021Deep} and capture the contextual relationships of the data\cite{Xiong2021Contextual}, respectively. Furthermore, by integrating LSTM networks with other components (such as graph neural networks\cite{Zhao2022Recons} and external memory\cite{Quan2020Recurrent}), learning 3D contexts and the temporal dynamics of multiple studies can accurately estimate 4D changes\cite{Zhang2020Spatio,Gessert2020Deep}, while exploiting the frame-level dependencies with LSTM (or the shot-level dependencies with graph convolutional networks)\cite{Zhao2022Recons} and remembering previous metaknowledge\cite{Santoro2016Meta} in the optimization of performance across similarly structured tasks can perform key-shot\cite{Zhao2022Recons} and one-shot learning\cite{Santoro2016Meta}, respectively. However, most memory mechanisms rely on weight-like storage (e.g., RNNs) or information-flow gating (e.g., LSTMs) rather than activity-based task-relevant information maintenance of working memory, which yields the best compressed transformative representation of dynamic environments for flexibility/generalizability across tasks\cite{Yoo2022How}.

Recently, deep reinforcement learning over working memory has pushed reward-maximizing artificial agents in interacting with their dynamic environments into learning to achieve working memory-like flexibility/generalizability across tasks\cite{Yoo2022How}. To exhibit the flexibility/generalizability of trainable working memory, a promising neural network architecture, i.e., working memory through attentional tagging, learns to flexibly control its memory representation in response to sensory stimuli in a biologically plausible fashion via reinforcement learning\cite{Kruijne2021Flexible}. This architecture can store arbitrary representations with its random, untrained encoding projections and has a built-in capacity to compute the degree of match between the representations in memory and incoming sensory information, such that it has raised the promising hope that only the organization of memory architecture potentially supports the learning of memory storage and retrieval to solve complex tasks with novel stimuli that it has never encountered before. However, such random feedforward encoding with built-in matching computation is not sufficient and generic enough for some challenging tasks with nonlinear combinations of overlapping heterogeneous inputs with complex interferences.   

\begin{figure*}[ht]
	\centering
	\includegraphics[width=0.8\textwidth]{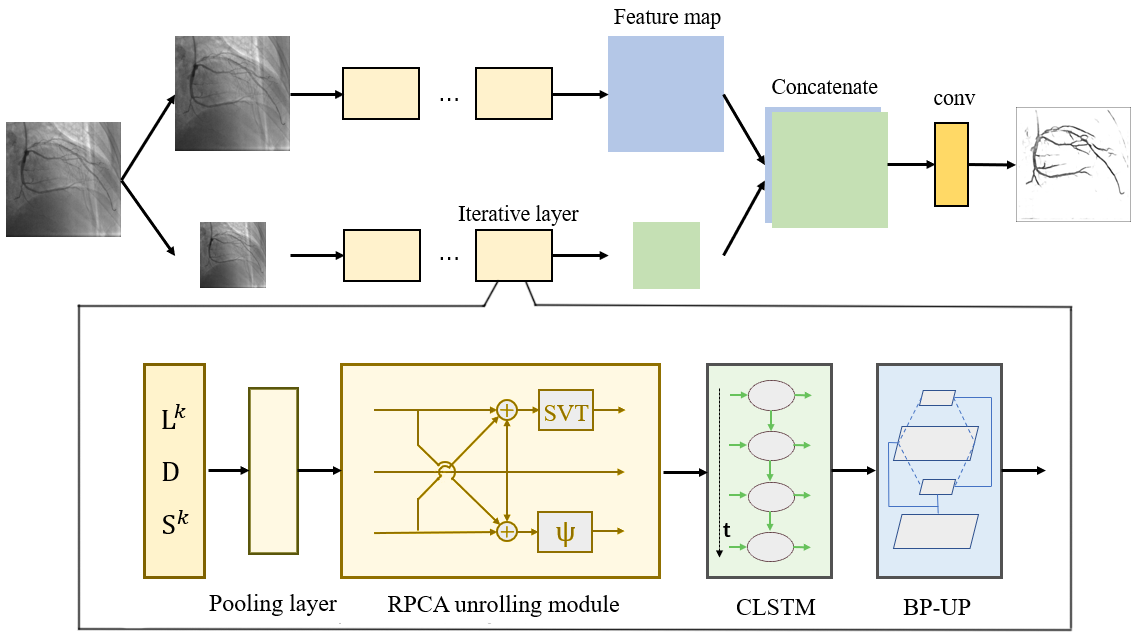}
	\caption{The overall architecture of the proposed msRPCA-PBNet. The network architecture are multiscale with each scale being composed of three parts: 1) Pooling layer downsamples the input patches for suppressing noises and motion interferences; 2) RPCA unrolling module globally separates moving contrast-filled vessels from the complex and noisy backgrounds in the XCA sequence; 3) CLSTM-based patch recurrent decomposition with backprojection/upsampling (BP-UP) superresolution module implements heterogeneous vessel superresolution and interference suppression.} 
	\label{Fig2} 
\end{figure*}

\section{Method}\label{sec:method}
Inspired by working memory that flexibly maximizes its efficiency and reduces interference via sparsification of memory representations and dynamically transforming representations to random orthogonal subspaces\cite{buschman2021balancing}, we propose dual-stage video decomposition via unfolded RPCA coupled with patch recurrent spatiotemporal decomposition to tackle the intricate overlapping and heterogeneous patterns of XCA sequences. Specifically, after globally decomposing an XCA sequence into foreground/background structures via RPCA unrolling, we further aggregate the decomposed patchwise structures via the CLSTM network to project them into spatiotemporally orthogonal subspaces, refining the underlying foreground/background patterns by suppressing noise and motion interferences. By prioritizing the more stable structured memory to constrain the less stable continuous/random memories of heterogeneous intensities and deformations of XCA vessels, this global-to-nonlocal transformative representation hierarchy is advantageous for working memory models to use the sparse/low-rank decomposition and patch recurrent orthogonal decomposition to smoothly regularize the encoding and retrieval of heterogeneous vessels from noisy and dynamic backgrounds. Similar representational transformations have also been explored in encoding and retrieval of short-term memory maintenance and long-term memory for episodic memory via intracranial EEG recordings with deep neural network models\cite{Liu2021Transformative}.

By building upon hierarchical WMIDPRUA with global-to-nonlocal transformative representations, the proposed mcRPCA-PBNet for dual-stage video decomposition is shown in Fig. \ref{Fig2} with base network architecture at each scale being composed of three parts: 1) a pooling layer downsamples the input patches for suppressing noise and motion interferences. This interference suppression is partially achieved due to not only increasing sparsity for reducing interference in neuronal information processing\cite{Ganguli2012Compressed} but also aggregating nonlocal patches after pooling operations that have been proven to provide some translation invariance in the convolutional neural networks\cite{LeCun1989BP}; 2) as a sensor layer of visual working memory, the multiscale patch-recurrent RPCA unrolling module implementing global video decomposition separates moving contrast-filled vessels from the complex and noisy backgrounds in the XCA sequence; 3) a CLSTM-based patch-recurrent backprojection/upsampling (BP-UP) superresolution module recurrently projects the decomposed vessel/background patches into spatiotemporally orthogonal subspaces for heterogeneous vessel retrieval and interference suppression.

\subsection{Multiscale RPCA Unrolling Networks}
Recently, foreground/background separation has become increasingly accepted as an accurate and robust strategy to solve the overlapping heterogenous problem for moving object extraction in visual surveillance and visual recognition applications. Specifically, by exploring sparse/low-rank structured representations for foreground/background structures, the RPCA model is defined to globally decompose input video sequence data into sparse foreground anomalies (moving contrast agents in XCA imaging) and low-rank components (slowly changing background structures):
\begin{equation}
	min\left \| L \right \| _{*} + \lambda \left \| S \right \| _{1}    s.t. D = L + S 
\end{equation} 
where $L$ is the low-rank component and $S$ is described as a sparse matrix with noise. $\left \| \cdot \right \|_*$ is the nuclear norm (which is the sum of its singular values), $\left \|\cdot\right \|_1$ is the $l_1$-norm regularization, and $\lambda$ is a regularizing parameter to control the extraction sensitivity to the number of sparse foreground components. The RPCA is formulated in a Lagrangian form as\cite{solomon2019deep}:
\begin{equation}
	min \frac{1}{2} \left \| M - H_{1}L - H_2S \right \| _{F}^2 + \lambda_1 \left \| L \right \| _{*} +  \lambda_2 \left \| S \right \| _{1,2}
\end{equation}
where $H_{1}$ and $H_{2}$ are the measurement matrices of $L$ and $S$ (in XCA images, $H_{1}$ = $H_{2}$ = $I$). $\left \| . \right \| _{1,2}$ is the mixed $l_{1,2}$-norm, and $\lambda_1$ and $\lambda_2$ are the regularizing parameters of $L$ and $S$, respectively.

By solving this equation via ISTA, we obtain an iteration solution with the iteration $k + 1$ being updated via
\begin{equation}
	L^{k + 1} = SVT_{\lambda_1 / L_f}{(I - \frac{1}{L_f}H_{1}^{H}H_1)L^k - H_{1}^{H}H_2S^k + H_1^HD}
\end{equation}
\begin{equation}
	S^{k + 1} = \psi_{\lambda_2 / L_f}{(I - \frac{1}{L_f}H_{2}^{H}H_2)L^k - H_{2}^{H}H_1S^k + H_2^HD}
\end{equation}
where $SVT_{\lambda_1/L_f}$ is the singular value thresholding operator, $\psi_{\lambda_2/L_f}$ is the soft-thresholding operator, and $L_f$ is the Lipschitz constant. After that, the above equations can be unrolled into convolutional layers by replacing coefficient matrices with convolutional kernels as follows:
\begin{equation}
	L^{k + 1} = SVT_{\lambda_1^k}{P_5^k*L^k+ P_3^k*S^k+P_1^k*D}
\end{equation}	
\begin{equation}
	S^{k + 1} = \psi_{\lambda_2 ^k}{P_6^k*S^k+ P_4^k*L^k+P_2^k*D}
\end{equation}
where $*$ denotes a convolutional operator. Here, convolutional layers $P_1^k$,..., $P_6^k$ and regularization parameters, as well as $\lambda_1^k$ and $\lambda_2^k$, are learned during the training process.

Then, we develop the RPCA unrolling network into a multiscale RPCA unrolling network, as shown in Fig. \ref{Fig2}. This multiscale RPCA unrolling is implemented with a patch-recurrent processing strategy (refer to \ref{sec3.2} for details). The input data are composed of data at the original resolution and scaled-down versions at different resolutions. When decomposing the input of different resolutions, the network can account for spatiotemporal correlation in different ranges, such that different feature information can be fully exploited. Finally, the multiscale outputs are adjusted to the original resolution and concatenated as input into a convolutional layer to obtain the final prediction result.

\subsection{Patch-recurrent Processing Strategy}\label{sec3.2}
Continuously moving contrast agents in different vessel branches with various shapes and thicknesses are sparsely distributed in the XCA sequence, such that the appearance and intensity of vessels vary nonlocally in XCA images. Therefore, globally decomposing XCA over entire images into foreground/background structures faces limitations in accurately restoring the heterogeneous vessel information while suppressing the complex interferences, in which a dynamic background with complex local deformations and mixed Poisson-Gaussian noises in XCA images largely affects the decomposition results. In clinical low-dose X-ray imaging, mapping raw X-ray images into a logarithmic domain is always required to compensate for the exponential attenuation of X-rays passing through the body, which results in  grey levels that are then linearly dependent on the matter thickness and density. Therefore, the mixed Poisson-Gaussian noises can be well modelled with an additive zero mean normal-distribution $\eta$ with signal dependent standard deviation ${{\sigma }_{\eta }}\left( S \right)$ as\cite{Hensel2006Robust}:
\begin{align}
	f(\eta ;S)=\frac{1}{{{\sigma }_{\eta }}\left( S \right)\sqrt{2\pi }}\exp \left( -\frac{{{\eta }^{2}}}{2{{\sigma }_{\eta }}{{\left( S \right)}^{2}}} \right)
\end{align}

This noise modelling results in the failure of global RPCA-based video decomposition over entire images for accurately extracting the heterogenous vessels, especially small vessel branches that are easily corrupted by signal-dependent noises and have large grey-level differences compared with the large vessels. Meanwhile, there is a great deal of feature variability between different XCA sequences acquired from heterogeneous environments. Global processing over entire XCA images may lead the neural networks to be biased in favour of majority features in different XCA sequences with class imbalance problems\cite{hao2020sequential}.  

In fact, XCA sequences lie in dynamic low-dimensional subspaces and are self-similar. The low dimensionality stems from the high correlation existing among the X-ray attenuation coefficients, and self-similarity is common in natural images and means that they contain many similar patches at different locations or scales in the images. This characteristic has been recently exploited by state-of-the-art patch-based image restoration methods. For mixed Poisson-Gaussian noise, the patched Gaussian mixed model is learned from a noisy image\cite{Irrera2016flexible,zhao2018detail,zhao2019texture}. Inspired by this observation, we divide the image into patches for multiscale vessel/background separation. The intensity of the vessel is then regarded as homogeneous, and the noise model follows a pure Gaussian distribution and is signal independent\cite{Irrera2016flexible}, such that accurately extracting heterogeneous vessels under mixed Poisson-Gaussian noises can be iteratively transformed into extracting homogeneous vessel patches under pure Gaussian noise.

\begin{figure*}[ht]
	\centering
	\includegraphics[width=0.76\textwidth]{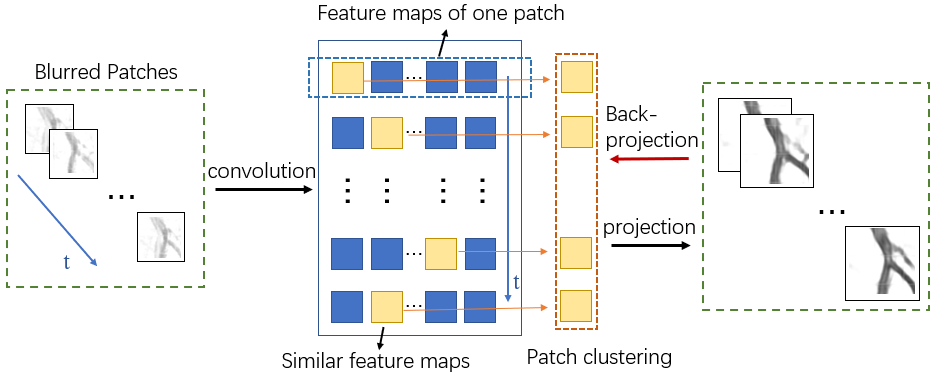}
	\caption{Patch clustering in CLSTM-based backprojection superresolution module. Input patches are formed into feature maps via convolution, where features of target vessels are extracted and the influence of slight motion between different frames is reduced. During the projection process of those feature maps, similar feature maps from different frames can be enhanced by each other. After that, backprojection is applied to increase the detail recovery of those feature maps.} 
	\label{Fig3} 
\end{figure*}

This work simultaneously exploits the self-similarity and dynamic low-dimensionality via spatiotemporally orthogonal decomposition (refer to \ref{sec3.3} for details) in XCA video sequences. We propose a sequential patch-recurrent processing strategy in a multiscale RPCA unrolling network to improve the capability of reducing the influences of complex interferences and vessel heterogeneity. Such a patch-recurrent processing strategy in a hierarchical way makes full use of long-range nonlocal similar structures in sparsely distributed small image patches to strengthen the ability to recognize moving objects and introduce much fewer special structures during the training step, which increases the sparsity of input features in the RPCA unrolling. Increasing feature sparsity is assumed to eliminate the probability of interference and enhance the robustness of the network in neuronal information processing\cite{Ganguli2012Compressed} and working memory tasks\cite{buschman2021balancing,panichello2019error}. Furthermore, by building upon an efficient translation invariance with theoretical guarantees of convolutional networks\cite{LeCun1989BP} used in CLSTM (refer to \ref{sec3.3}), patch-recurrent processing can aggregate nonlocal similar patches to suppress background motion interferences in vessel/background separation. Therefore, exploring both self-similarity and low dimensionality enables the XCA video decomposition to be formulated with respect to the patch-recurrent random representation, thus greatly improving the decomposition performance and reducing the computational complexity during processing.

\subsection{CLSTM Backprojection Superresolution}\label{sec3.3}
To refine the vessel candidates from vessel-like background artefacts and complex interferences, we proposed a CLSTM-based backprojection superresolution module after the RPCA unrolling network. The CLSTM-based superresolution module is inspired by our previous work\cite{Qin2022RPCAUNet}, in which CLSTM can store and select spatiotemporal correlation features in the memory cell, including sequential intensity and structure information flow in the previous frames. Specifically, the temporal relations between frames can be extracted via the LSTM structure, while the spatial structure is kept intact by convolution operations for the recurrent gating mechanism\cite{shi2015convolutional}. Because XCA vessels appeared as spatially distributed moving agents in previous frames and then gradually disappeared into the dynamic and noisy backgrounds in subsequent frames, we decoupled the spatial and temporal dimensions of the XCA video sequence via CLSTM to refine vessel representations from overlapping artefacts as well as noises and motion interferences. In addition, because spatial and temporal spaces are orthogonal and independent of each other, this CLSTM-based decomposition motivates us to decompose spatial and temporal feature representations first and then project this decomposition into random patch-recurrent spaces. Different from our previous work\cite{Qin2022RPCAUNet}, we are inspired by working memory to implement transformative representations by integrating global-to-nonlocal video decomposition into patch recurrent backprojection\cite{haris2021deep} for heterogeneous vessel superresolution and motion interference suppression. 

Long short-term memory was first proposed by Hochreiter and Schmidhuber\cite{hochreiter1997long} for natural language processing and then expanded into CLSTM by adding convolutional structures\cite{shi2015convolutional} into a recurrent architecture for image/video processing. The main idea of CLSTM is the introduction of a hidden memory cell, which can enable the network to propagate the temporal information of previous data. The CLSTM replaces fully connected layers in LSTM with convolutional layers. The formula of CLSTM is as follows:
\begin{align}
	& f^{t} = \sigma (W_f \ast x^t + U_f \ast h^{t-1}+ V_f \ast c^{t-1} + b_f), \notag \\
	& i^{t} = \sigma (W_i \ast x^t + U_i \ast h^{t-1} + V_i \ast c^{t-1} +b_i), \notag \\
	& o^{t} = \sigma (W_o \ast x^t + U_o \ast h^{t-1} + V_o \ast c^{t-1} + b_o),  \notag \\
	& c^t = f^{t} \circ c^{t-1} + 
	i^{t} \circ tanh(W_{c} \ast x^t + U_{c} \ast h^{t-1} + b_c), \notag \\
	& h^t = o^{t} \circ  tanh(c^t)
\end{align}
where $\ast$ is the convolutional operator, $\circ$ is the Hadamard product, $x^t$ denotes the input, $c^t$ denotes the memory cell that can store the temporal information of previous input data, and $i^t$, $f^t$ and $o^t$ denote the controlling gates where information in memory cells can be written, cleared and propagated, respectively. $h^t$ is the final output that is determined by the current input $x^t$ and the hierarchical information stored in $c^t$.

Patched vessel structures exhibit slight movements in different frames due to heartbeats. This movement effect can be solved by the translational-invariance convolution layer in CLSTM with patch clustering, as shown in Fig. \ref{Fig3}. Input patches are formed into feature maps through a convolutional layer, and then these feature maps are projected into high-resolution space via deconvolution. Such unstructured random projection performed by recurrent deconvolution is able to deblur the image and enhance high-frequency information \cite{zhao2021sparse}. During the process of deconvolution, similar feature maps can achieve patch clustering. Specifically, among those feature maps from different frames, where the influence of slight motion between different frames is reduced, many similar feature maps appear. During the recurrent deconvolution, these similar feature maps can be clustered and enhanced by each other. Therefore, extracting features from multiframe patches via CLSTM can achieve adaptive patch clustering to reduce the motion and noise interferences while preserving the vessel features.

However, detailed information, e.g., small vessel branches and distal vessels, is easily blurred. We apply a backprojection\cite{haris2021deep} mechanism with convolution and deconvolution operations to recover the detailed vessel information. Specifically, feature maps can be downsampled back into low-resolution space by convolution. Then, the difference between the low-resolution feature maps and origin feature maps can be used to guide the final deconvolution operation. Such backprojection mechanism (see Fig. \ref{Fig3}) embedded in the CLSTM-based patch recurrent decomposition recurrently learns the spatiotemporal correlations between multiscale image patches with different image resolutions to effectively minimize the vessel superresolution reconstruction error. This backprojection mechanism has been successfully applied in single image superresolution where iterative up- and down-projection modules are used with the reconstruction error being iteratively calculated\cite{haris2021deep}.

In the proposed network, a backprojection module is applied to upproject the feature maps $h^{t}$ into a high-resolution space. The backprojection module is defined as follows:
\begin{align}
	& H_{0}^{t} = (h^{t} \ast p_t)\uparrow  _s, 
	\notag \\
	&h_{0}^{t} = (H_{0}^{t} \ast g_t)\downarrow  _s,
	\notag \\
	&e_t^l = h_{0}^{t} - h^{t},
	\notag \\
	&H^t_1 = (e_t^l \ast q_t) \uparrow  _s, 
	\notag \\
	&H^t = H_0^t + H^t_1
\end{align}
where $\ast$ is the convolution operator, $p_t$ and $q_t$ are the deconvolutional layers, $g_t$ is the convolutional layers, $\uparrow _s$ and $\downarrow _s$ represent up-sampling and down-sampling operator, respectively. The backprojection module projects the input feature map $h^{t}$ into an intermediate high resolution map $H_{0}^{t}$ via a deconvolution layer. Then, the intermediate high-resolution map is downprojected back into a low-resolution map $h_{0}^{t}$. It is obvious that if the high-resolution map is close to the ground truth, the low-resolution $h_0^t$ and $h^{t}$ should be infinitely close as well. Therefore, the residual between $h_0^t$ and $h^{t}$ , which is defined as $e_{t}^{l}$, can be used to guide the final high resolution output. Finally, the high resolution output map {$H^{t}$ is obtained by summing the intermediate high resolution map $H_{0}^{t}$ and the high resolution residual map $H_{1}^{t}$ (obtained by deconvolution operation on $e_{t}^{l}$}), which enables better preservation of original feature details.

\subsection{Automatically Generated Weakly-supervised Labelling}
We aim to extract both the geometric features and image grey values of XCA vessels, which are unrealistic for manual labelling. Therefore, an automatic vessel labelling for weakly supervised learning is implemented using a vessel region background completion method (VRBC)\cite{qin2019accurate}, which is the only method available to accurately and automatically recover vessel intensity information while rarely introducing background structures. Specifically, VRBC first extracts vessel structures from complex and noisy backgrounds by RPCA-based vessel extraction. An accurate binary mask of the vessel is finely generated via Radon-like feature filtering with spatially adaptive thresholding. Subsequently, vessel-masked background regions are recovered to complete background layers by implementing tensor completion with the spatiotemporal consistency of whole background regions. Finally, the layers containing vessels' greyscale values can be accurately extracted by subtracting the completed background layers from the overall XCA images.
\begin{figure*}[htbp]
	\centering 
	\includegraphics[width=1\textwidth]{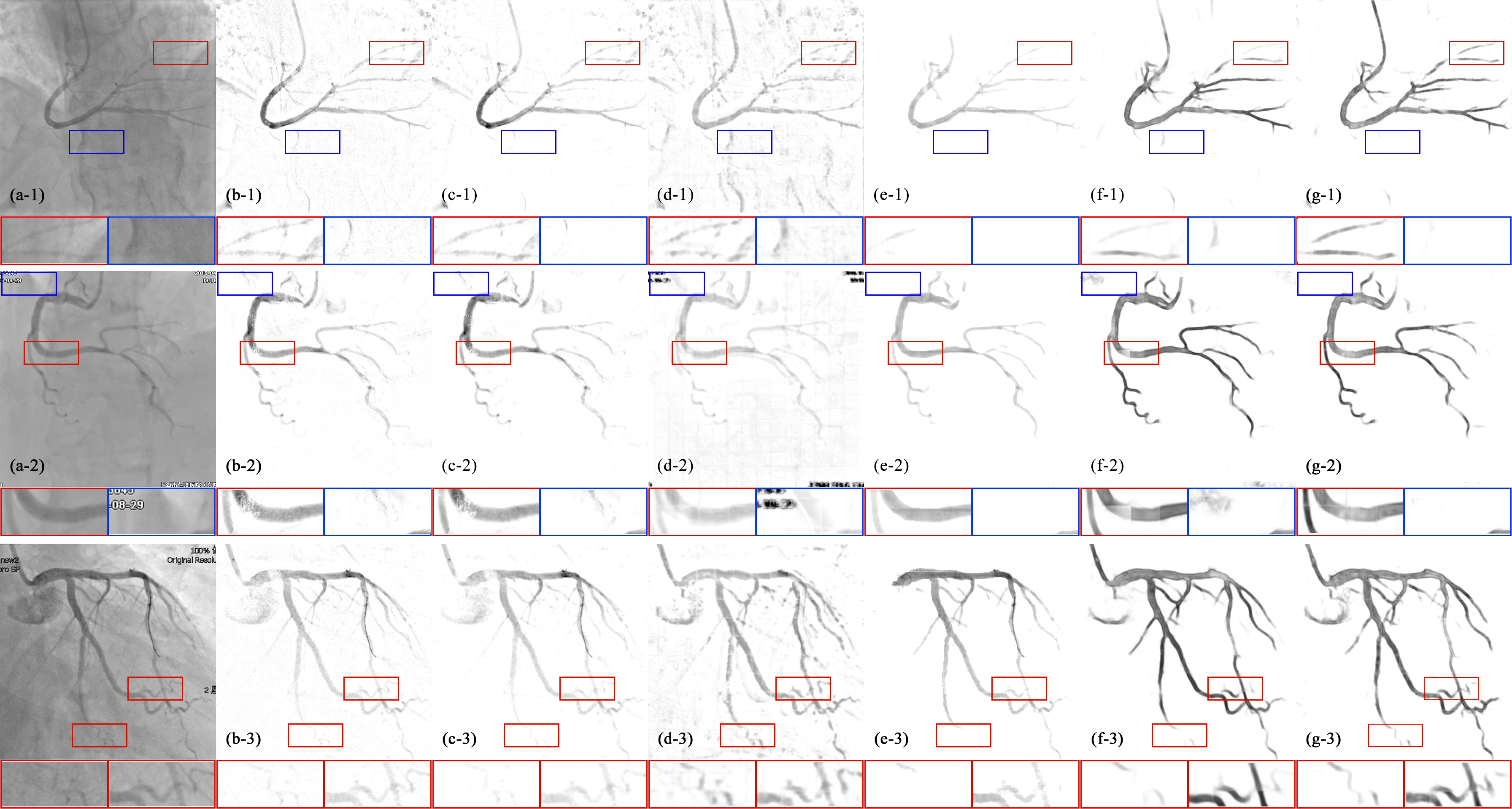}	
	\caption{XCA vessel extraction results. (a) Original XCA image. (b) MoG-RPCA\cite{zhao2014robust}. (c) MCR-RPCA\cite{jin2017extracting}. (d) CORONA\cite{solomon2019deep}. (e) VRBC\cite{qin2019accurate}. (f) RPCA-UNet\cite{Qin2022RPCAUNet}. (g) msRPCA-PBNet. Red box regions show the small vessel detection performance and blue box regions show the interference suppression performance}	
	\label{Fig4} 
\end{figure*}

\section{Experimental Results}\label{sec:Experimental results}
\subsection{Data}
We have collected 43 real clinical XCA sequences from Renji Hospital affiliated to Shanghai Jiao Tong University School of Medicine\cite{Qin2022RPCAUNet}. To evaluate our method's learning performance in heterogeneous environments, these sequences are accquired from different patients and different imaging machines, including X-ray angiography systems from Philips and Siemens. Therefore, these sequences have various characteristics in aspect of image quality, grey level and noise distribution. The length of each sequence ranges from 30 to 140 frames. The resolution of each frame is 512 $\times$ 512 pixels, with 8 bits per pixel. To optimize the quality of training data, we have removed the violently disturbed frames, which solely have severe background interferences without real XCA vessels, from the dataset used in our previous work\cite{Qin2022RPCAUNet}. Finally, nearly 1800 frames are selected for our experiments. The weakly-supervised labels of training pairs for the proposed mcRPCA-PBNet are automatically generated by the VRBC method\cite{qin2019accurate}, which can recover vessel intensity information while rarely introducing background artefacts. The ground truth vessel masks for the quantitative evaluation of vessel extraction in \ref{sec4.5} and the evaluation of vessel segmentation in \ref{sec4.6} are obtained via manually annotation by three experts. 

\subsection{Experiment Settings}
The proposed moving contrast-filled vessel extraction networks\footnote{The source code will be available at  \url{https://github.com/Binjie-Qin/msRPCA-PBNet.}} consists of 4 iterative layers and three scales, including the original resolution and the resolutions after downsampling 2 times. In each iterative layer, the RPCA unrolling module contains 6 convolutional layers. The first two iterative layers use convolutional kernels of size = 5 with stride = 1, padding = 2 and a bias, and the other two layers use convolutional kernels of size = 3 with stride = 1, padding = 1 and a bias. The long short-term memory backprojection superresolution module contains a CLSTM feature extraction layer, a backprojection/upsampling layer and an output layer. The CLSTM feature extraction layer uses convolutional kernels of size = 3, channels = 64, stride = 1, and padding = 1. The backprojection/upsampling layer uses convolutional kernels of size = 6, channels = 64, stride = 2, and padding = 2. The output layer uses convolutional kernels of size = 3, stride = 1, and padding = 1.

We choose the adaptive moment estimation (ADAM) optimizer with a learning rate of 0.0001 and mean square errors (MSE) as the loss function. The XCA sequences are randomly divided into training, validation and test datasets at a ratio of approximately 0.6:0.2:0.2. The XCA sequences are divided into 64 $\times$ 64 $\times$ 20 patches with a certain overlap (50$\%$ between two neighbouring patches). Totally, 20000 patches are generated and used in the experimental evaluation.

\subsection{Comparison Methods}
We used several state-of-the-art RPCA-based methods for comparison, including MoG-RPCA \cite{zhao2014robust}, our previous MCR-RPCA \cite{jin2017extracting}, CORONA \cite{solomon2019deep} and VRBC\cite{qin2019accurate}. Additionally, our previous RPCA-UNet\cite{Qin2022RPCAUNet} is also used for comparison in our experiment. After vessel extraction, the binary vessel mask can be segmented by a traditional threshold method such as Otsu \cite{otsu1979threshold} to achieve a vessel segmentation result for vesssel segmentation performance evaluation. Therefore, we can evaluate the performance of vessel segmentation by comparing our method with advanced segmentation algorithms such as Frangi's \cite{frangi1998multiscale}, Coye's \cite{coye2015novel}, SVS-net \cite{hao2020sequential} and CS$^2$-net \cite{mou2021cs2}. Generally, the overall performance of all methods is consistent with and slightly better than that demonstrated in our previous work\cite{Qin2022RPCAUNet}, since the dataset adopted in this work is refined for performance optimization, i.e., the violently disturbed frames have been removed from the training dataset used in our previous work\cite{Qin2022RPCAUNet}.

\subsection{Visual Evaluation of Vessel Extraction}
The moving contrast-filled vessel extraction results are shown in Fig. \ref{Fig4}, where the regions in red boxes show the small vessel extraction performance and the regions in blue boxes show the interference suppression performance. Traditional RPCA-based methods achieve good performance in major vessel extraction. The major vessel components can be clearly distinguished from the background. However, the contrast between distal vessels and the background is relatively small, and there are still many background artefacts in the final results, which further reduces the visual effect of distal vessels. Although this phenomenon has been improved in the MCR-RPCA method, such performance still has much room for improvement. The VRBC-based method has made significant improvement in the suppression of background artefacts, and the obtained extraction results contain almost no components other than target vessels. However, break points exist in the vessel extraction results, especially in the positions where the contrast between vessels and the background is low.

\begin{figure*}[htbp]
	\centering	
	\centering 
	\includegraphics[width=1\textwidth]{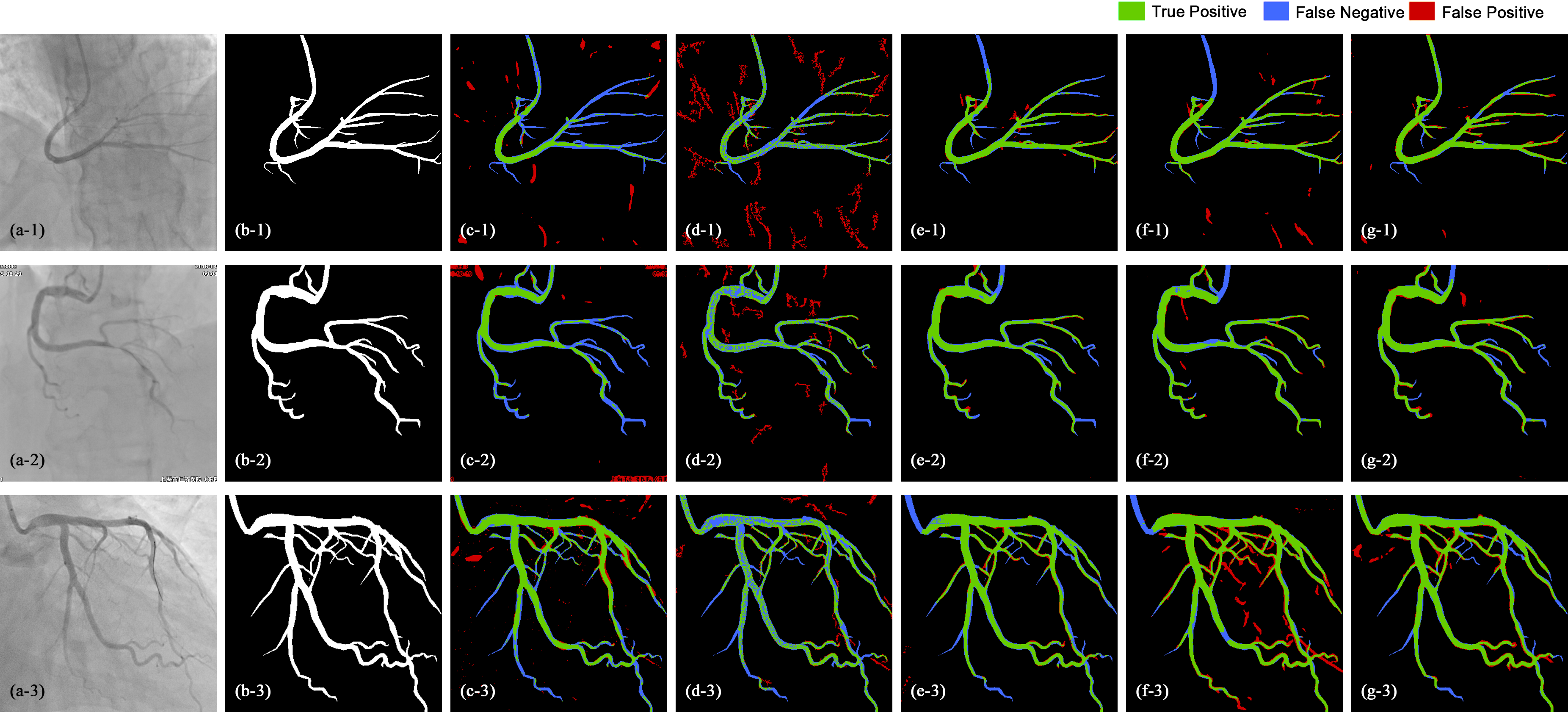}	
	\caption{XCA vessel segmentation results. Pixels labelled with green, blue, and red represent true positive pixels, false negative pixels, and false positive pixels, respectively. (a) Original XCA image. (b) Ground-truth vessel mask. (c) Frangi's\cite{frangi1998multiscale}. (d) Coye's\cite{coye2015novel}. (e) SVS-net\cite{hao2020sequential}. (f) CS$^2$-net\cite{mou2021cs2}. (g) msRPCA-PBNet.}
	\label{Fig5} 
\end{figure*}

Compared to these methods, RPCA unrolling-based RPCA-UNet and msRPCA-PBNet significantly improve the vessel extraction performance since the extracted vessel tree structure is more complete and clear. The msRPCA-PBNet method presents more pure results with fewer background impurities compared to the previous RPCA-UNet. Moreover, the proposed method performs better in vessel detail extraction, especially for the distal vessels with low contrast. For example, in the first row of Fig. \ref{Fig4}, red box region has quite low contrast between distal vessels and background, VRBC and RPCA-UNet can hardly extract such vessel details clearly. Traditional RPCA-based methods can extract some of them while introduce many background artefacts. However, msRPCA-PBNet can extract distal vessels exactly without introducing artefacts and noises. In blue box regions, the edges of spine structure have similar intensities and shapes to XCA vessels, RPCA-UNet and traditional RPCA-based methods extract them wrongly while msRPCA-PBNet remove the interference of such vessel-like structures. Furthermore, the regions in red and blue boxes of Fig. \ref{Fig4}(f-2) and Fig. \ref{Fig4}(g-2) clearly show that msRPCA-PBNet outperforms RPCA-UNet in recovering continuous heterogeneity of vessel information with simultaneously removing block and blur effects as well as residual interferences.

\subsection{Quantitative Evaluation of Vessel Extraction}\label{sec4.5}
The vessel visibility can be quantitatively evaluated by the contrast-to-noise ratio (CNR) \cite{solomon2019deep}. A larger CNR means better vessel visibility. The CNR can be calculated as follows:
\begin{equation}
	CNR =  \frac{\left | \mu_{V} - \mu_{B}  \right |}{ \sqrt{\sigma_B^{2} + \sigma_V^{2}}  } 	
\end{equation}
where $\mu_{V}$ and $\mu_{B}$ are the pixel intensity means in the vessel and background regions, respectively, and $\sigma_V$ and $\sigma_B$ are the standard deviations of the pixel intensity values in the vessel regions and background regions, respectively.

\begin{table}\normalsize
	\caption{The average CNR values (mean value $\pm$ standard deviation)}
	\label{TabI}
	\setlength{\tabcolsep}{3.5mm}{
		\begin{tabular}{lll}			
			\hline			
			Method& Global CNR& Local CNR\\
			\hline
			MoG-RPCA\cite{zhao2014robust}& 1.372 $\pm$ 0.173& 1.346 $\pm$ 0.192\\
			MCR-RPCA\cite{jin2017extracting}& 1.333 $\pm$ 0.166& 1.286 $\pm$ 0.180\\
			CORONA\cite{solomon2019deep}& 1.280 $\pm$ 0.223& 1.287 $\pm$ 0.288\\
			VRBC\cite{qin2019accurate}& 1.390  $\pm$ 0.262& 1.356 $\pm$ 0.262\\
			RPCA-UNet\cite{Qin2022RPCAUNet}& 1.765 $\pm$ 0.284& 1.633 $\pm$ 0.206\\
			Ours& \textbf{1.795 $\pm$ 0.276}& \textbf{1.645 $\pm$ 0.257}\\
			\hline
	\end{tabular}}
	\label{TabI}
\end{table}

To comprehensively evaluate vessel visibility, we consider the quality of the global image and surrounding regions of vessels, which has the greatest impact on human observation. Therefore, global and local background regions are defined to cover all the image regions except the vessel regions and the 7-pixel-wide neighbourhood regions surrounding the vessel regions, respectively\cite{qin2019accurate}. The CNR calculation results are shown in Table \ref{TabI}. The results show that the proposed method achieves the highest global and local CNRs, indicating that the visibility of the extraction result produced by the proposed network achieves great improvement both globally and locally. 

\subsection{Visual Evaluation of Vessel Segmentation}\label{sec4.6}
The vessel segmentation results are shown in Fig. \ref{Fig5}. To better show the difference between the segmentation results and the manually annotated ground truth, we use different colours to label the different types of pixels, in which green pixels are the true positive pixels that are correctly classified as vessels, blue pixels are false negative pixels that are vessel pixels but wrongly classified as backgrounds, red pixels represent false positive pixels that are wrongly classified as vessels but actually belonging to the backgrounds. The segmentation results show that Frangi's method\cite{frangi1998multiscale} can segment the major vessel regions whose intensity is easily distinguishable but cannot detect some heterogeneous vessel branches. The Coye's method\cite{coye2015novel} can detect vessels with different intensities; however, it introduces many background components and is badly corrupted by the strong noises. 

The supervised deep learning-based SVS-net\cite{hao2020sequential} and CS$^2$-net\cite{mou2021cs2} achieve better performance. The SVS-net segments the major vessels accurately without introducing background impurity. However, it fails to detect most of the small vessel branches. In contrast, CS$^2$-net exhibits a good ability to detect distal vessels while apparently introducing relatively more background artefacts. Additionally, discontinuity may occur in the major vessel regions. Our msRPCA-PBNet achieves the best performance, as it can segment the relatively complete vessel tree in an automatic weakly-supervised way. Although the segmentation of distal vessels is slightly weaker than that of CS$^2$-net, it can segment the major vessel and most distal vessels stably while hardly introducing background artefacts.

\begin{table} \normalsize
	\caption{Means and Standard Deviations of the DR, P, and F metrics on twelve XCA images}
	\label{2}
	\begin{center}	
		\setlength{\tabcolsep}{0.76mm}{
			\begin{tabular}{lccc}				
				\hline				
				Method& Detection Rate& Precision & F-measure\\
				\hline				
				Frangi's\cite{frangi1998multiscale}& 0.556 $\pm$ 0.254& 0.773 $\pm$ 0.205&  0.647 $\pm$ 0.214    \\
				Coye's\cite{coye2015novel}& 0.678 $\pm$ 0.047& 0.717 $\pm$ 0.185& 0.662 $\pm$ 0.095      \\
				SVS-net\cite{hao2020sequential}& 0.727 $\pm$ 0.098& 0.912 $\pm$ 0.058&  0.806 $\pm$ 0.071   \\
				CS$^2$-net\cite{mou2021cs2}& 0.789 $\pm$ 0.098& 0.867 $\pm$ 0.074&  0.821 $\pm$ 0.064   \\
				Ours& \textbf{0.818 $\pm$ 0.078}& 0.865 $\pm$ 0.065& \textbf{0.838 $\pm$ 0.055}      \\
				\hline
		\end{tabular}}
	\end{center} 
	\label{2} 
\end{table}

\begin{figure*}[htbp]
	\centering
	\subfigure{
		\begin{minipage}[t]{0.17\linewidth}
			\centering
			\includegraphics[width=1.3in]{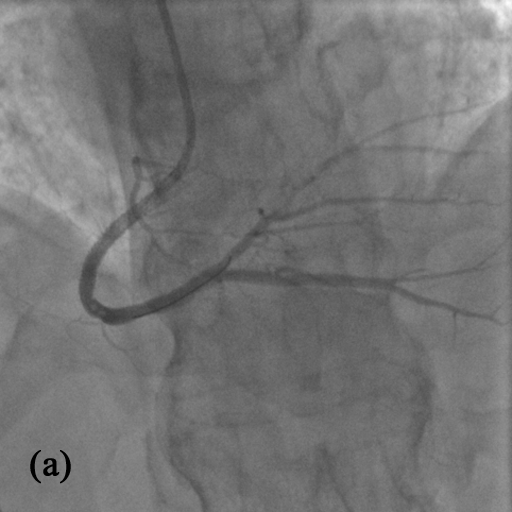}
		\end{minipage}%
	}%
	\subfigure{
		\begin{minipage}[t]{0.17\linewidth}
			\centering
			\includegraphics[width=1.3in]{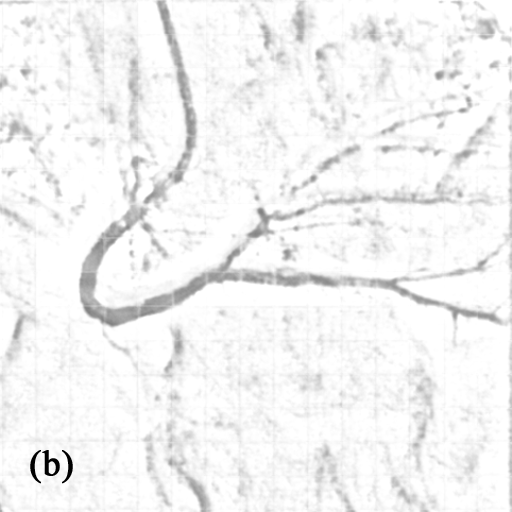}
		\end{minipage}%
	}%
	\subfigure{
		\begin{minipage}[t]{0.17\linewidth}
			\centering
			\includegraphics[width=1.3in]{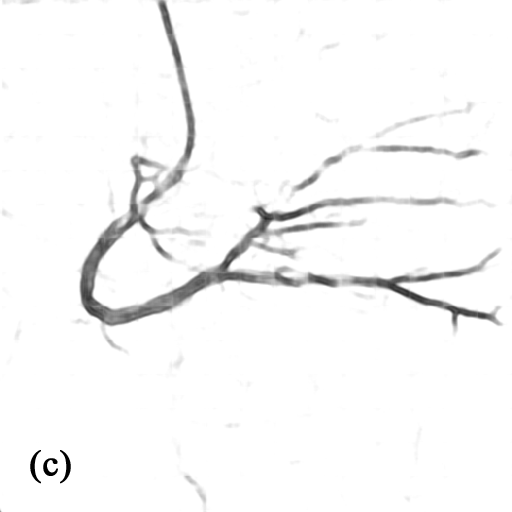}
		\end{minipage}
	}%
	\subfigure{
		\begin{minipage}[t]{0.17\linewidth}
			\centering
			\includegraphics[width=1.3in]{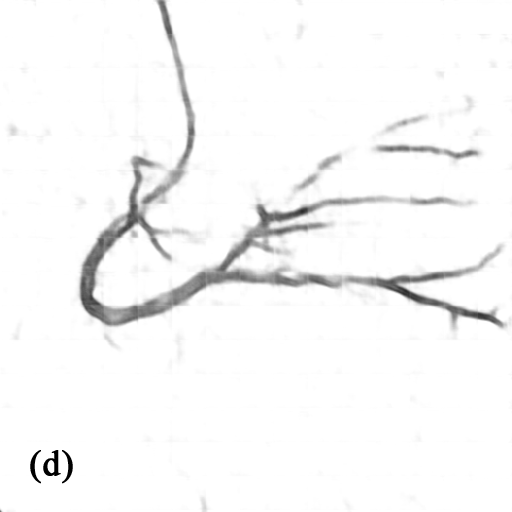}
		\end{minipage}
	}%
	\subfigure{
		\begin{minipage}[t]{0.17\linewidth}
			\centering
			\includegraphics[width=1.3in]{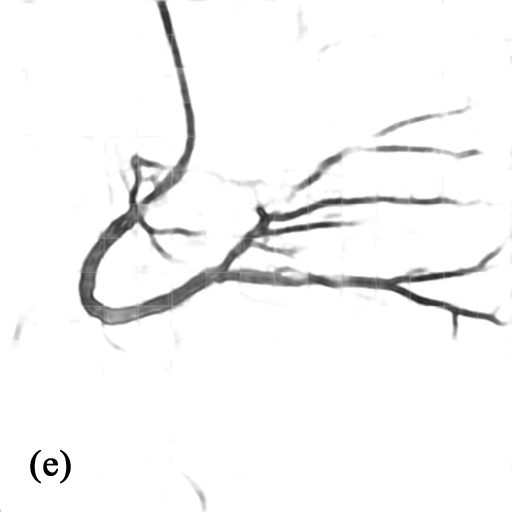}
		\end{minipage}
	}%
	\caption{Results of the ablation study. (a) Original XCA image. (b) RPCA unrolling network. (c) RPCA unrolling network with a backprojection module. (d) Multiscale RPCA unrolling network with a backprojection module. (e) Multiscale RPCA unrolling network with CLSTM backprojection module}
	\label{Fig6}
\end{figure*}

\subsection{Quantitative Evaluation of Vessel Segmentation}
The performance of the segmentation results can be evaluated by the detection rate (DR), precision (P) and F-measure (F). The DR represents the ratio of vessel pixels that were successfully classified to the total vessel pixels in the ground truth. The precision represents the ratio of correctly classified vessel pixels to the total vessel pixels in the segmentation result. F measure depends on both DR and P measures, which is an evaluation indicator that reflects the comprehensive performance of the segmentation result. These metrics can be calculated as follows:
\begin{equation}
	DR = \frac{TP}{TP + FN} , P = \frac{TP}{TP + FP} , F = \frac{2\times DR \times P}{DR + P} 
\end{equation}
where TP (true positive) represents the number of foreground pixels that are correctly classified, FP (false positive) represents the number of background pixels that are incorrectly classified as foreground, TN (true negative) represents the number of background pixels that are correctly classified and FN (false negative) represents the number of foreground pixels that are incorrectly classified as background. 

The DR, P, and F measures of the proposed msRPCA-PBNet and other state-of-the-art segmentation methods are displayed in Table \ref{2}. The proposed msRPCA-PBNet achieves the highest DR and F measure in the experiment. In the evaluation of the P value, the proposed method produces a lower value than SVS-net because the number of false positive pixels is small in SVS-net's result, while many small blood vessels are ignored. Therefore, the DR of SVS-net is lower than that of the proposed method. In general, msRPCA-PBNet produces a better comprehensive performance.
\begin{table} \normalsize
	\caption{Training time and runtime of networks in ablation study}
	\label{3}
	\begin{center}	
		\setlength{\tabcolsep}{0.76mm}{
			\begin{tabular}{lccc}				
				\hline				
				Method& Training time(h)& Runtime(s)\\
				\hline				
				RPCA Unrolling\cite{solomon2019deep}& 53.333 & 0.608      \\
				RPCA Unrolling + BP& 26.042& 0.352      \\
				Multiscale RPCA Unrolling + BP& 36.153& 0.499     \\
				Ours& 70.611& 1.275    \\
				\hline
		\end{tabular}}
	\end{center} 
	\label{3} 
\end{table} 

\subsection{Ablation Study}
To investigate the role of each module in msRPCA-PBNet, we designed an ablation experiment that compares the following networks: an RPCA unrolling network, an RPCA unrolling network with a backprojection module, a multiscale RPCA unrolling network with a backprojection module, and a multiscale RPCA unrolling network with a CLSTM backprojection module. The results of the ablation experiments are shown in Fig. \ref{Fig6}. 

The RPCA unrolling network can complete the preliminary extraction of the moving vessel layer, while many vessel-like structures can also be captured into the vessel layer due to their slight movement. The network embedded with the backprojection module significantly improves the extraction result and eliminates most of the background impurities. However, there are still some background impurities in the surrounding areas of vessels, which interferes with the visual observation. With the addition of a multiscale mechanism, the network can obtain a larger range of spatiotemporal information from the input patches of different scales, eliminating some background artefacts that do not have connectivity in the image, but it also leads to the ignorance of some small vessels with low contrast. The proposed msRPCA-PBNet that integrates the multiscale mechanism and the long short-term memory backprojection module can handle this problem by making full use of the long- and short-range spatiotemporal information stored in the memory cell. Therefore, it achieves the best extraction result, where the vessel tree is relatively complete and few background impurities are introduced.

Moreover, Table \ref{3} shows the training time and runtime of these networks in ablation study. The training time and runtime of original RPCA unrolling networks is higher than those of the networks embedded with the backprojection module, since the number of layers for the original RPCA unrolling network is chosen as 10 for a relatively good performance while the layer numbers of other improved networks only need to be set to 4 for a better performance. Generally, the running time of these networks can basically meet the clinical needs.

\section{Conclusion and Discussion}\label{sec:Conclusion}
Inspired by a flexible working memory model, we proposed dual-stage deep video decomposition networks with transformative representation hierarchy between multiscale patch recurrent RPCA unrolling networks and a CLSTM-based backprojection superresolution module, which can accurately extract the structure and intensity information of heterogeneous moving objects while suppressing complex noises and motion interferences in the challenging XCA sequence. Specifically, the iterative RPCA unrolling networks serves as a sensory layer to perceive the sparse/low-rank structured represenations of global foreground/background decomposition, while the CLSTM-based backprojection acts as the role of a control layer in working memory to refine and project these perceived foreground candidates into the unstructured random representations of nonlocal patch recurrent decomposition in spatiotemporally orthogonal subspaces, recognizing sequential moving foregournd items from the overlapping interferences. These dual-stage decompositions have a supplementary effect to each other and efficiently capture the important discriminative features of subspace information for heterogeneous object reconstruction. Most importantly, the global-to-nonlocal transformative representations have been ignored in current RPCA-based video decomposition to remove overlapping interferences from complex and dynamic backgrounds. 

Furthermore, being developed from our previous RPCA-UNet\cite{Qin2022RPCAUNet}, whose weakly supervised learning performance and generalization ability trained by automatically-genenrated coarse labels are better than that trained by manually-labelled fine labels, msRPCA-PBNet also can remove the cost spent on manually labelling data and further improve the generalizability of deep video decomposition simultaneously since the patch recurrent dual-stage decomposition with transformative representations leads to fewer moving object representations that need to be learned by the networks. Generally, the underlying WMIDPRUA for the proposed msRPCA-PBNet enables the deep video decomposition networks to achieve ad hoc interpretability and computational efficiency as well as flexibility/generalizability in extracting moving objects against complex and dynamic background interferences.


%



\ifCLASSOPTIONcompsoc
  \section*{Acknowledgments}
\else
  \section*{Acknowledgment}
\fi
The authors would like to thank all the cited authors for providing the source codes used in this work and the anonymous reviewers for their valuable comments on the manuscript.

\ifCLASSOPTIONcaptionsoff
  \newpage
\fi



%



\bibliographystyle{IEEEtr}
\bibliography{bare_adv}

%








\end{document}